\documentclass[]{Liebert_Author}

\title{AcousTac: Tactile sensing with acoustic resonance for electronics-free soft skin} 

\author{Monica S. Li,$^{1}$ Hannah S. Stuart,$^{1\ast}$\\
{$^{1}$Department of Mechanical Engineering, University of California Berkeley,}\\
{Berkeley, CA, USA}\\
{$^\ast$To whom correspondence should be addressed;}\\
{E-mail:  hstuart@berkeley.edu.}
}

\begin{document} 

\maketitle 


\begin{abstract}
Sound is a rich information medium that transmits through air; people communicate through speech and can even discern material through tapping and listening. To capture frequencies in the human hearing range, commercial microphones typically have a sampling rate of over 40kHz. These accessible acoustic technologies are not yet widely adopted for the explicit purpose of giving robots a sense of touch. 
Some researchers have used sound to sense tactile information, both monitoring ambient soundscape and with embedded speakers and microphones to measure sounds within structures. However, these options commonly do not provide a direct measure of steady state force, or require electronics integrated somewhere near the contact location. 
In this work, we present AcousTac, an acoustic tactile sensor for electronics-free force sensitive soft skin. Compliant silicone caps and plastic tubes compose the resonant chambers that emit pneumatic-driven sound measurable with a conventional off-board microphone. The resulting frequency changes depend on the external loads on the compliant end caps. Non-idealized vibration of the caps results in a boundary condition transition when first making contact, initially producing nonmonotonic signals. We characterize two solutions -- adding a distal hole and mass to the cap -- resulting in monotonic and nonhysteretic force readings with this technology. We can tune each AcousTac taxel to specific force and frequency ranges, based on geometric parameters, including tube length and end-cap geometry and thus uniquely sense each taxel simultaneously in an array. We demonstrate AcousTac's functionality on two robotic systems: a 4-taxel array and a 3-taxel astrictive gripper. Simple to implement with off-the-shelf parts, AcousTac is a promising concept for force sensing on soft robotic surfaces, especially in situations where electronics near the contact are not suitable. Equipping robots with tactile sensing and soft skin provides them with a sense of touch and the ability to safely interact with their surroundings. 
\end{abstract}

%%%%%%%%%%%%%%%%%%%%%%%%%%%%%%%%%%%%%%%%%%
\section{Introduction}
\label{sec:intro}

Soft robotic surfaces and skins generate inherent safety useful in human robot physical interaction and interaction with fragile objects \cite{pang_2021}. Soft systems and mechanisms physically embody the ability to adapt to dynamic and unexpected loads through passive compliance \cite{laschi_2016, rus_2015}. As a result, soft robots also exhibit durability in harsh environments \cite{tolley_soro_2014}.
This motivates the development of soft, compliant coverings for otherwise rigid robots, for both robot and environmental safety. For example, the CoboSkin includes variable stiffness inflatable units to tune impact dynamics \cite{pang2020coboskin}. A recent review by Niiyama points out how soft skin is essential for humanoid robot application \cite{niiyama2022soft}. Yet, these hardware solutions are slow to translate to industrial settings, as compared with sensing and software based collision avoidance \cite{simoes2022designing}. The design of soft skins for various robotic applications therefore remains a relevant area for ongoing development. 

Artificial skin can also provide robots with a sense of touch that enables adaptive control for dexterous manipulation tasks \cite{dahiya_sense_2010}. Composition of sensitive skins relies on a wide variety of materials, including those with resistive, piezoelectric, and magnetic properties; the history and breadth of tactile sensor design is reviewed in chapters, reports, and books \cite{cutkosky2016force, howe_1993, dahiya2013robotic, wan_2017}. Skin with channels filled with liquid-phase gallium-indium alloy is one method for generating sensitive skin, in which overlaid channel patterns are individually sensitive to unidirectional strain and contact pressure \cite{park_2012}. Numerous other works utilize bulk resistivity and embed cavities of conductive material in soft structures, where contact forces change the shape and therefore resistance \cite{truby_soft_2018, giffney_2016, koivikko_2018}. All of these works describe a direct electronics-based measurement of the contact point. 
For skin made of multiple materials, challenges arise regarding the durability of traditional electronics, integration of soft and hard conductive materials without failure from stress concentrations or fatigue at connectors, and long-term life and reliability of experimental new materials at the contact point \cite{roberts2021soft,dahiya2013robotic}. Early development of self-healing soft tactile skin combats some of these durability issues \cite{markvicka_2018}.  

Removing electronics and delicate materials from the contact point altogether represents another approach to generating mechanically resilient touch sensation. 
Various options place an intermediary material between physical stimuli and electric transducers to achieve such isolation. 
For example, recent work by Sun, et al. generated a camera-based structure, as a successor technology to GelSight \cite{yuan2017gelsight}, that shifts the camera to the base of the finger; it provides an indirect measure of force while improving mechanical durability of the appendage \cite{sun2022soft}. Pneumatics and hydraulics are another avenue for transmitting data throughout soft structures for remote sensing. 
Monitoring the pressure in a closed fluidic system provides rich tactile sensing information \cite{truby_2022, fishel_2012}. 
In open fluidic tactile sensors, flow rate monitoring discerns object presence and surface characteristics for underwater \cite{nadeau_2020} and in air \cite{huh_2021} applications. In these examples, transducers are physically in-line with flow or pressure differential tubes routed away from the contact location. 
Vibrations transmitted through structures provide yet another method specialized for dynamic tactile sensing \cite{cutkosky2014dynamic}. Resonant frequency-based contact detection informs robotic reaching and grasping, such as work by Backus, et al. that used accelerometers in compliant fingers \cite{backus_2014}. Such vibration measurements require the vibrations to transmit effectively through the robotic structure itself. 

Beyond the use of sight, pressure, or structural vibrations to measure contact from a distance, sound is another medium through which to measure touch. Sound, generated through either the active or passive emission, can be measured with a microphone. 
The sounds of two surfaces sliding across each other tell us about relative movement and the material type, shown for rigid-rigid contact in Dornfeld and Handy \cite{dornfeld_ae_1987}. Coincidental noise during manipulation can even be used to provide force feedback during teleoperation \cite{shi2022touching}. These signals change with material properties, and do not provide a direct measure of either static or dynamic contact forces. 
In an active sensing application, a receiver monitors the ultrasound signal transmitted through an air cavity, that changes with stress in the surrounding compliant material \cite{shinoda_1997}. Researchers embed a microphone in a soft pneumatic finger and later add an embedded speaker that outputs a frequency sweep to detect contact with high spatial resolution \cite{wall_2022,zoeller_2018}. Acoustic vibrations between robots assist in shared communication and coordination \cite{drew2021acoustic}. These active acoustic examples achieve more specific measurands than in passive sound generation, however the emitter/receiver pairs are embedded or located in close proximity to the contact. 

In the present work, we present AcousTac, an active, pneumatically-driven, acoustic tactile sensing method suitable for force measurements using a generic microphone located outside the skin. The idea is that robots, with a regulated air supply and a built-in microphone on the body, can include this sensing modality without the addition of new electronics. 
Instead of using an electronic sound emitter, we take inspiration from wind instruments to produce sound at the skin. Air-driven resonance is an ancient concept, with musical instruments emerging thousands of years ago.\footnote{The flute may even date back 60,000 years \cite{bower_1998}.} In recent years, researchers have used simulations to model and design musical instruments \cite{fabre_2012,fletcher_1979}. 
We design a tactile array using air-driven pipe resonance, because it is based on a simple theoretical model and uncomplicated to fabricate. 
This design has no electronic components at the contact area, removing the need to robustly integrate fragile components and route wires in soft skin. 
In our prior work, we generated such resonance in tubes along the length of a soft finger in order to detect the pose and fingertip contact, and estimate force with a fully-rigid probe \cite{li_2022}. The present work represents the first study looking at this modality for force sensitive soft skin, and includes new observations about how resonant modes and sound of a soft structure change when interacting with an object. 

%%%%%%%%%%%%%%%%%%%%%%%%%%%%%%%%%%%%%%%%%%
\subsection{Overview}
In Section \ref{sec:design}, we describe the theoretical basis for AcousTac design, which includes models of taxel length and resonant frequency, relationships of taxel deformation with force, and proposed alterations of boundary conditions. The taxel implementation is presented in Section \ref{sec:methods}, along with a description of the experimental and signal processing methods used to characterize the sensor. Results from experimental characterizations of different single taxels in Section \ref{sec:results} reveal the effect of tube length and end cap modifications on the range and sensitivity of AcousTac. We demonstrate a system of AcousTac sensors in a 4-taxel array and a 3-taxel astrictive gripper in Section \ref{sec:demo}. Design guidelines, additional observations, and limitations and future work are described in Section \ref{sec:disc}. Section \ref{sec:concl} summarizes the impact of this work in the broader context.

%%%%%%%%%%%%%%%%%%%%%%%%%%%%%%%%%%%%%%%%%%
\section{Taxel design}
\label{sec:design}

In the following sections, we describe the theory behind the design parameters examined in this work. We present tradeoffs in generating acoustic resonance in structures that respond to variable force ranges and with different compliant boundary conditions. 

%%%%%%%%%%%%%%%%%%%%%%%%%%%%%%%%%%%%%%%%%%
\subsection{Resonant frequency and tube length}
\begin{figure*}
\centering
    \includegraphics[width=0.9\linewidth]{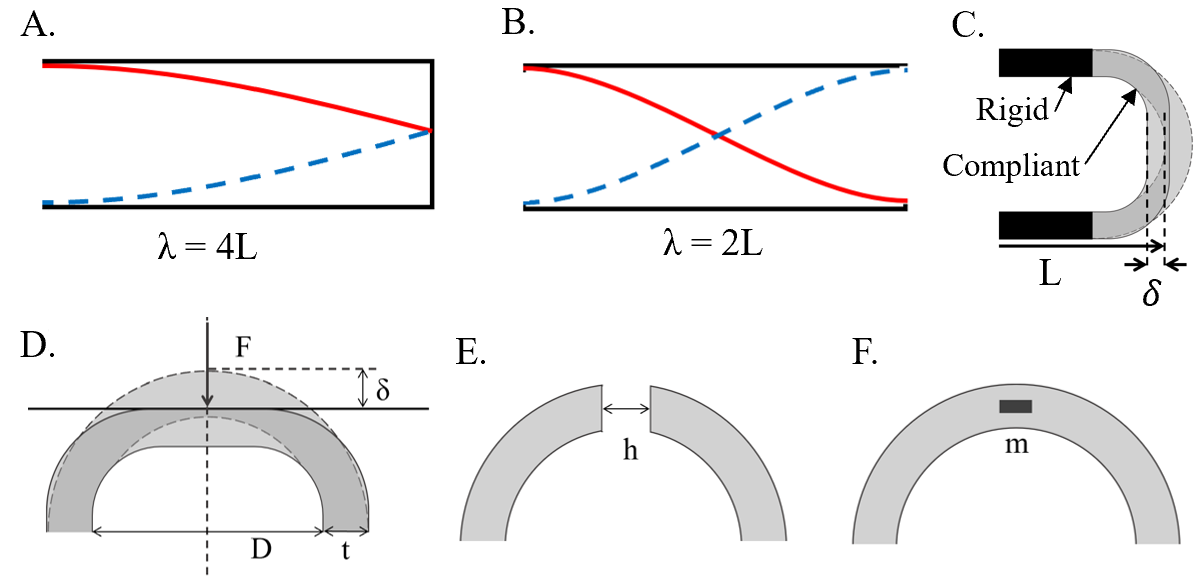}
	\caption{A. Open-closed boundary condition, where L is one-quarter of the wavelength $\lambda$. B. Open-open boundary condition. L is half wavelength. We imagine the boundary condition as open-damped, where the damped end is in between open and closed. C. Schematic denoting rigid tube length $L$ and compliant cap deformation $\delta$ that changes the total length of the taxel. 
 D. Schematic of hemispherical cap. A normal force $F$ deforms the cap by distance $\delta$. This deformation decreases the total length of the taxel and increases the resonant frequency. The inner diameter of the cap $d$ is kept constant. We test varying wall thickness $t$ and adding E. a hole with diameter $h$ and F. mass $m$. 
 }
	\label{fig:theory}
\end{figure*}

In one dimensional (1D) theory, a tube that is open on one end and closed on the other has a fundamental resonant wavelength, $\lambda$, four times the tube length, $L$ (Fig. \ref{fig:theory}A). Frequency, $f$, is related to wavelength, $\lambda$, by $\lambda = c/f$, where $c$ is the the speed of sound. Resonant frequency, $f_{oc}$, of an open-closed tube is therefore 
\begin{equation}
    f_{oc} = \frac{c}{4L}.
    \label{eqn:1doc}
\end{equation}
For a tube open on both ends (Fig. \ref{fig:theory}B), the resonant wavelength is twice the tube length, such that the resulting frequency, $f_{oo}$, is
\begin{equation}
    f_{oo} = \frac{c}{2L}.
    \label{eqn:1doo}
\end{equation}
For both the open-open and open-closed boundary condition, frequency is inversely proportional to length. The frequency will be higher for a smaller sensors with shorter tubes. We propose to fit experimental data to a modification of Eqn. \ref{eqn:1doc} in this case, with the constants $b_i$ such that
\begin{equation}
    f = \frac{b_1}{b_2 \: L} + b_3.
    \label{eqn:bfit}
\end{equation} 
In the case of a tube with a soft end cap at one end, this boundary condition appears to lie between the open and closed idealized cases, as the soft material partially damps and reflects vibrations, still resulting in resonance. When external forces deform the soft end-cap  (Fig. \ref{fig:theory}C), the length of the tube changes; the frequency also changes. Linear approximation of Eqn. \ref{eqn:1doc} shows that the shorter the tube length $L$, the greater frequency change $\Delta f$ for a change in length $\delta$, where $L$ is the unloaded length. As such, shorter tubes are more sensitive to end cap deformations. For a linear approximation of the open-closed tube, change in frequency relates to length as
\begin{equation}
    \Delta f = -\frac{c}{4} \frac{\delta}{L^2}, \: \: \Delta f \propto  \frac{\delta}{L^2}
    \label{eqn:Delta_fL}
\end{equation}

The above 1D theory assumes a thin tube, $L >> D$, where $D$ is the inner diameter of the tube. In practice, taxel tubes have a non-zero diameter and a particular inlet geometry which makes the measured resonant frequency deviate from theory. Regardless, the 1D model serves as a useful design tool to predict first-order trends. Additional factors known to contribute to resonant frequency, are kept constant for this study. For example, as flow rate increases, higher harmonics dominate the resonance; this work assumes the fundamental frequency mode across all designs.

%%%%%%%%%%%%%%%%%%%%%%%%%%%%%%%%%%%%%%%%%%
\subsection{Force on compliant hemispherical shell}
Deformations of the end-cap alter tube length and result from externally applied normal compressive forces. Frequency is therefore a measure of force. Designers can tune force sensing range and resolution by varying end-cap stiffness. Both material selection and geometry alter the force-deformation response. Generally, the stiffer the cap, the greater the force range and the lower the resolution and sensitivity. 
The Hertzian contact theory model for small deformations of a compliant rubber hemisphere relates force $F$ to displacement $\delta$ as
\begin{equation}
    F \propto \delta ^{3/2}
    \label{eqn:def}
\end{equation} 
In this work, the end-cap is a hemispherical shell with a wall thickness $t$ and inner diameter $D$ that experiences large deformations (Fig. \ref{fig:theory}D). 
We therefore propose an empirically fit approximation of Eqn. \ref{eqn:def}, for each cap design 
\begin{equation}
    F = \beta_1 \delta ^{\beta_2} + \beta_3.
    \label{eqn:betafit}
\end{equation}

%%%%%%%%%%%%%%%%%%%%%%%%%%%%%%%%%%%%%%%%%%
\subsection{Altering end-cap boundary conditions} 

Due to the physical properties of the soft compliant end-cap, the unloaded cap behaves somewhere in between an open and closed boundary condition, and we refer to this as \textit{damped}. In this damped boundary condition, the cap is a flexible, oscillating surface, resulting in a frequency higher than in the closed condition and lower than in the open condition. When a rigid object contacts the cap, the boundary condition undergoes a transition from damped to closed, lowering the resonant frequency. 

This will occur even without gross deformations or substantial changes in tube length $L$. However, this transition between different boundary conditions occurs across a range of light contact forces greater than zero, until there is no relative vibration between the object and cap surface. For forces higher than this boundary condition transition phase, gross deformations reduce length and increase frequency. 
We therefore expect a nonmonotonic sensor signal, with a minimum value under light contact forces. 

In this work, we propose and investigate two design features to mitigate or remove this resonance property, and produce reliable measures of force. Specifically, we explore the introduction of a hole or mass at the tip of the end-cap.

\subsubsection{Hole}
As in Fig. \ref{fig:theory}E, we add a hole of diameter $h$ to the center of the compliant cap. Adding a hole to the cap makes an unloaded cap behave more similarly to an open boundary. When an object occludes the hole and loads the cap, the end then transitions to the closed boundary condition. 
For a given tube length, the resonant frequency of the open-open boundary condition is twice the frequency of the open-closed boundary condition. Higher flow excites higher resonant frequency, and we generally expect increasing flow rate to excite higher modes. Therefore, selecting a flow rate that actuates closed tube resonance but not the open tube's creates a change in amplitude at the transition between no contact and contact. By coupling both amplitude and frequency measurements, contact force is uniquely determined, overcoming the nonmonotonic force-frequency relationship in the transition region. 
In addition to flow, some geometries, including the edge-orifice shape and inner tube diameter, are more conducive to certain resonance modes than others. 

\subsubsection{Mass}
As in Fig. \ref{fig:theory}F, we add a mass $m$ to the cap to bring it towards the closed boundary condition, even without contact. The mass increases the inertia of the compliant surface and further dampens the pressure oscillations. Sufficient mass removes the nonmonotonic behavior of the frequency, such that the boundary condition does not change upon contact.

%%%%%%%%%%%%%%%%%%%%%%%%%%%%%%%%%%%%%%%%%%
\section{Methods}
\label{sec:methods}

\subsection{Taxel implementation and test parameters}

\begin{figure*}[ht]
\centering
	\includegraphics[width=1\linewidth]{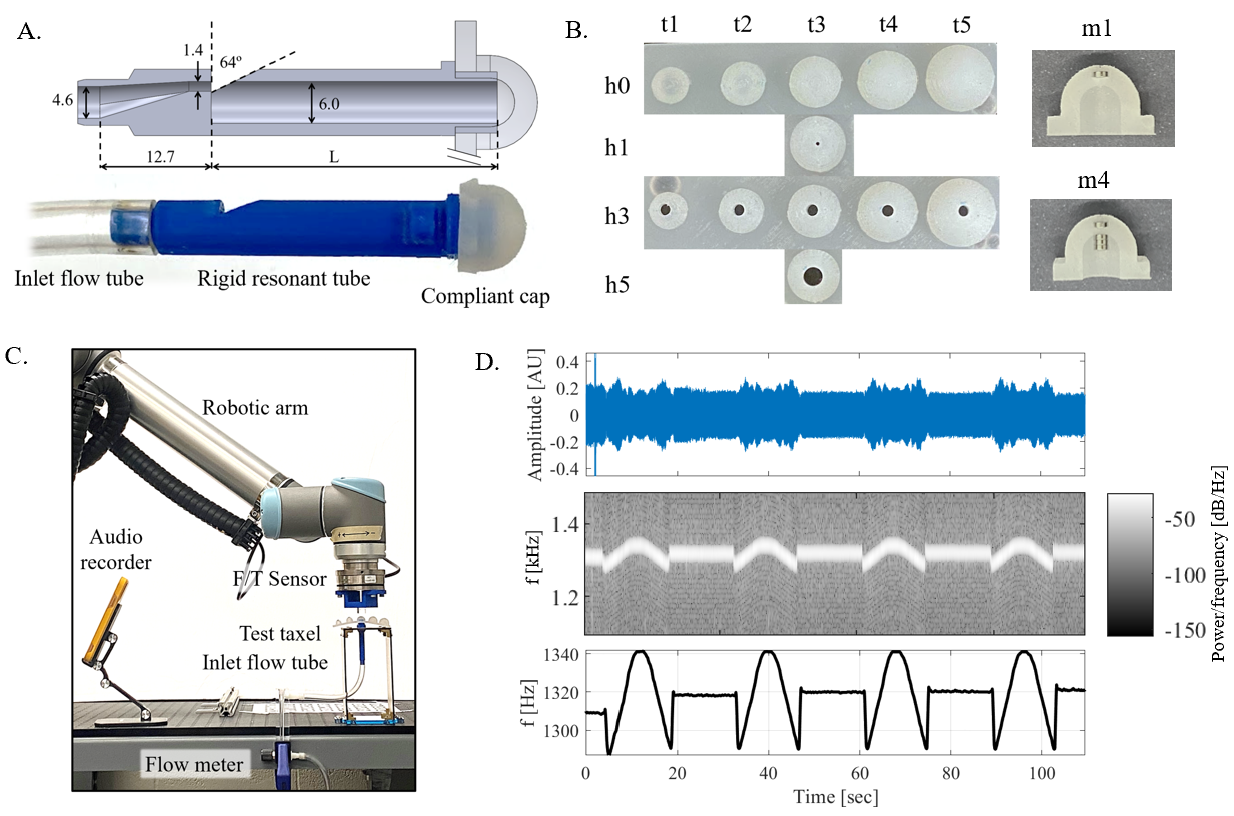}
	\caption{A. Taxel schematic L = 41 + 6n [mm]. All dimensions in mm, except for edge-orifice angle in degrees. Photo of fabricated taxel, analogous to above. Flow enters the resonant tube from the left. The hemisphere on the right is compliant and adhered to the rigid structure. B. Photo of the caps tested. We varied wall thickness $t$ and hole diameter $h$. The number indicates the wall thickness or hole diameter in millimeters. C. Photo of experimental setup for taxel characterization. D. Example data from processing the raw audio signal to spectrogram for frequency, then picking out the maximum amplitude frequency. 
	}
	\label{fig:setup}
\end{figure*}

A rigid tube and compliant end-cap compose each taxel. We individually fabricate these components and assemble them together into a single taxel, which is then rigidly mounted to an acrylic plate for testing. The rigid tube is a specified length $L$ (Fig. \ref{fig:setup}A), additive manufactured with PLA (Makergear M3-1D 3D Printer, nozzle diameter 0.35\,mm). The length of the resonant tube varies between 41 and 65\,mm, in increments of 6\,mm. These 6\,mm intervals ensure the resonant frequencies of the different taxels do not overlap. The inner diameter is 6\,mm. The smallest cross section is 1.4\,mm, located immediately before the edge-orifice, the angled cutout of the tube wall. The inlet and edge-orifice geometry are consistent across all taxels. 

The end-cap is the soft element in the taxel. We fabricate and test a range of cap thicknesses $t$, hole diameters $h$, and added mass $m$ (Fig. \ref{fig:setup}B) to assess the effect of these parameters. We cast silicone (Smooth-On Dragon Skin 30) in a two-part mold to make the caps. All caps have the same 7\,mm inner diameter hemisphere, regardless of thickness, and a 5\,mm long cylindrical cavity, also 7\,mm in diameter, that stretches over the open end of the rigid tube. 

To assess taxel stiffness and force range, the wall thickness $t$ of the caps ranges from 1 to 5\,mm (t1 through t5), at 1\,mm increments. To test the effect of hole size, we compare caps with 3\,mm wall thickness, across 0, 1, 3 and 5\,mm hole sizes. A complete set of thicknesses (t1 through t5) are tested for both without (h0) and with a 3\,mm hole (h3) to test coupling between these parameters. The caps with holes are cast with dowel pins of specified diameter in the negative mold. 
We add cylindrical magnets to the compliant caps to assess the addition of mass. Each magnet has a mass of 50\,mg, a diameter of 2\,mm, and a height of just less than 1mm. We test caps of 1, 3, and 5\,mm wall thickness, and vary the number of magnets between 1 and 4 (m1 through m4). This results in an added mass of up to 200\,mg. For the 1\,mm thickness cap, we adhere 1 magnet to the outer surface with glue. For 3\,mm and 5\,mm thickness, the first magnet is embedded into the cap during the casting process. 
We increase mass by adding more magnets to the inner surface of the cap, which utilizes their magnetic attraction to attach them. 

In this study, we characterize all the different end-cap designs using one resonant tube length $L=59$\,mm.
When comparing the different rigid tube lengths, we use a single end-cap design with 3\,mm wall thickness, and no hole nor mass (t3h0). 

\subsection{Taxel characterization test setup}
We characterize each taxel individually using the experimental setup shown in Fig. \ref{fig:setup}C. The taxel is fixed to the tabletop. Wall compressed air flows through flexible 8.3\,mm (3/8") tubing and a flow meter (analog 1-10\,LPM) before entering the resonant tube, which is kept between 4 to 5\,L/min. A robot arm (UR-10, Universal Robots) and 6DOF wrist force/torque sensor (Axia80, ATI) cyclically palpate the taxel, moving at 0.5\,mm/s until at least 5\,N of normal force. We increase the tested force range with cap thickness. 
Robot Operating System (ROS) software records force and position at 2.0\,kHz and 3.3\,kHz, respectively. A smartphone (iPhone 11) located approximately 0.5\,m away records sound at a sampling rate of 44.1\,kHz. Ambient lab noise is approximately 65\,dB, less than the approximately 90\,dB sound level when a taxel is active. 

%%%%%%%%%%%%%%%%%%%%%%%%%%%%%%%%%%%%
\subsection{Audio processing}

Fig. \ref{fig:setup}D shows an example of the audio signal and processing outcome for four palpations of the t3h0 end-cap.
We perform data processing using the pipeline detailed in Li, et al \cite{li_2022}. We extract the resonant frequency from the raw amplitude data at 25\,Hz, using the built-in MATLAB functions \textit{spectrogram()} and \textit{tfridge()} and binning frequency in 2.5\,Hz intervals. For caps with holes, we are also interested in the amplitude of the audio signal. 
We generate an envelope of the raw audio by finding the max absolute amplitude value for every 22.7\,ms (1000 data points) then taking a smooth average for every 113\,ms (5000 data points). The amplitude is then downsampled to 25\,Hz to match the sampling rate of extracted frequency. 

Characterization tests without a hole h0t1-h0t5 or with a mass m1-m4 all consist of at least 3 palpation cycles performed within 80\,sec each. In the amplitude thresholding tests when a hole is present, h3t1-h3t5 and h1t3-h5t3, we only report points above the set threshold, which is approximately 12\,sec of data over at least 2 palpation cycles. The specific amplitude thresholds in each hole case are individually selected as the minimum value to avoid a nonmonotonic relationship between force and frequency in that design's data set.  

In later demonstrations with systems of 3 and 4 taxels emitting sound simultaneously, we assume that the achievable frequency range of each taxel does not overlap with any other. This is achieved by assigning a unique tube length to each taxel. 
We process data within the expected frequency range for each taxel separately. 

%%%%%%%%%%%%%%%%%%%%%%%%%%%%%%%%%%%%%%%%%%
\section{Results: Taxel characterization }
\label{sec:results} 

\subsection{Tube length, $L$}
\label{sec:results:tubelength}

\begin{figure}[h]
\centering
	\includegraphics[width=1.0\linewidth]{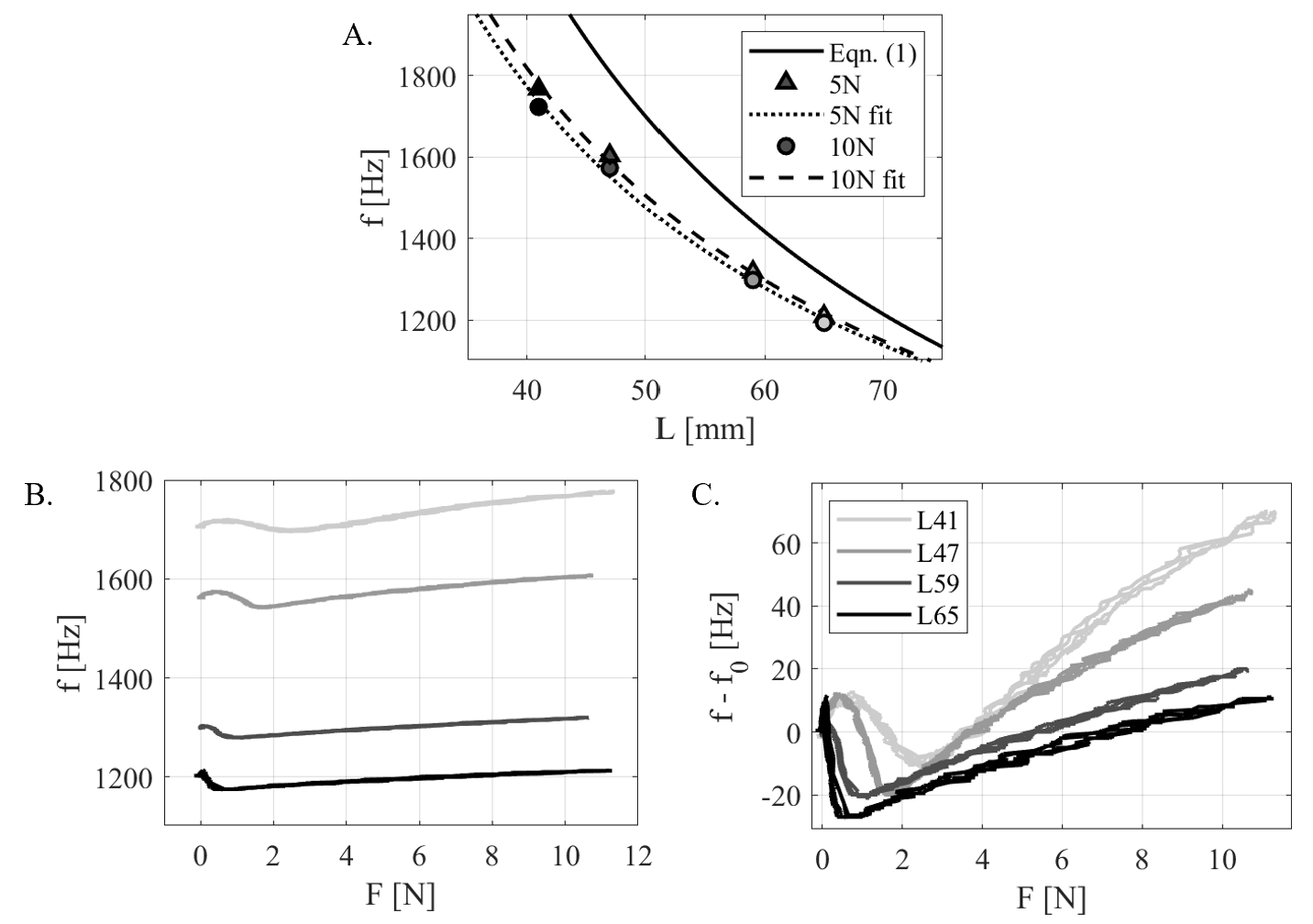}
	\caption{Calibration curves for taxels with varying tube length $L$ with cap t3h0. A. Tube length and frequency $f$. The 1D theoretical model (Eqn. \ref{eqn:1doc}) is shown in a solid black line. Experimental data for resonant frequency loaded at 5\,N (triangle) and 10\,N (circle). This data is fitted with Eqn. \ref{eqn:bfit}, shown with the dotted black lines. B. Resonant frequency with force $F$ for all four tested tube lengths. C. Net frequency change from unloaded condition, $f-f_0$, plotted with force. 
	}
	\label{fig:L}
\end{figure}

In Fig. \ref{fig:L}A, we plot the measured frequencies for the four rigid tube lengths from two normal force loads: 5\,N and 10\,N. We also show the 1D theory for an open-closed cavity from Eqn. \ref{eqn:1doc}. The measured frequency is consistent with the theory, in which longer tubes exhibit lower resonant frequency. 
Table \ref{tab:bi} presents fitting constants from Eqn. \ref{eqn:bfit}. The end-cap geometry is a negligible constant length offset and is not included in our calculations.

\begin{table}[h]
\centering
\begin{threeparttable}
\caption{Fitting constants for Eqn. \ref{eqn:bfit} plotted in Fig. \ref{fig:L}A. 
\label{tab:bi}}
\setlength{\tabcolsep}{12pt}%
\begin{tabular}{@{}llll}
\toprule
Fit constants & $f_{oc}$  & 5N & 10N  \\
\midrule
 b$_1$ & 340   & 290    & 280     \\
b$_2$ & 4         & 4.6    & 4.8     \\
b$_3$ & 0         & 260    & 290   \\
\bottomrule
\end{tabular}
\end{threeparttable}
\end{table}

Fig. \ref{fig:L}B shows the full characterization curves for frequency as a function of force, with loads ranging from zero to greater than 10\,N. By design, the frequency ranges do not overlap each other to streamline later signal processing when multiple taxels emit sound at the same time. 
In Fig. \ref{fig:L}C, we plot the same data but now as the net change in frequency; measured frequency $f$ is subtracted by unloaded frequency $f_0$ to inspect the calibration curve shape. 
All four taxels demonstrate a nonmonotonic frequency relationship, with both local maxima and minima observed at forces less than 3\,N. As force increases from initial contact, the frequency increases, then decreases, then increases again. This behavior is likely due to the boundary condition changing from damped to closed as the force increases. 
After the boundary condition transitions, at forces greater than 3\,N, the shorter length taxels with higher unloaded frequency have steeper slopes. This is consistent with Eqn. \ref{eqn:Delta_fL}. We do not observe any hysteresis in the signal for these taxel lengths.

\subsection{Cap wall thickness, $t$}

\begin{figure*}[h!]
\centering
	\includegraphics[width=0.95\linewidth]{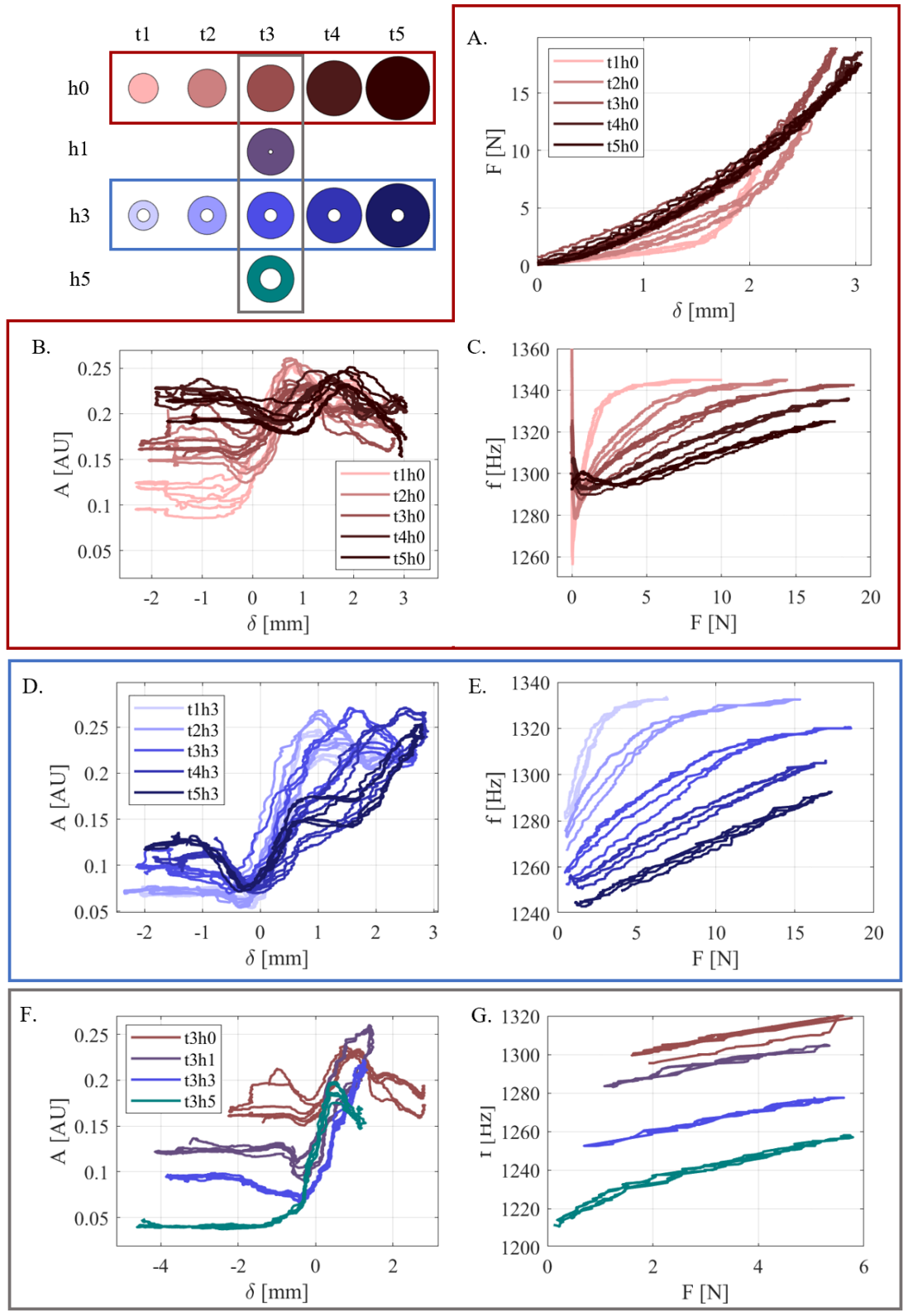}
	\caption{Characterization of cap wall thickness $t$ and hole diameter $h$, and their relationships with  force $F$, displacement $\delta$, audio amplitude $A$, and frequency $f$. A color coded schematic is used as a legend of tested cap parameters. 
 }
	\label{fig:tihj}
\end{figure*}

We plot cap deformation versus force in Fig. \ref{fig:tihj}A for caps with no hole (t1h0-t5h0). In the tested displacements, thicker caps provide a more linear force displacement curve, which is steeper than thinner caps at lower displacements. 
The softer, thinner caps have a linear region at low forces, then start to saturate. Table \ref{tab:betai1} shows the fitting constants for these caps using Eqn. \ref{eqn:betafit}.

\begin{table}[h]
\centering
\begin{threeparttable}
\caption{Fitting constants for Eqn. \ref{eqn:betafit} for the data plotted in Fig. \ref{fig:tihj}A. 
\label{tab:betai1}}
\setlength{\tabcolsep}{12pt}%
\begin{tabular}{@{}llll}
\toprule
Thickness, $t$ [mm] & F$_\text{max}$ [N]  & $\beta_1$ & $\beta_2$  \\
\midrule
 1 &   2  &  0.94    &    1.36   \\
2 &   6        &   1.56   &   1.77    \\
3 &     11      &   2.66   &   1.64  \\
4 &     15     &   2.65   &   1.68  \\
5 &     15      &   3.52   &   1.50  \\
\bottomrule
\end{tabular}
\end{threeparttable}
\end{table}

Fig. \ref{fig:tihj}B shows how the amplitude of sound produced by the taxel changes with deformation. A negative value of $\delta$ denotes the distance between contact surface and top of the cap, e.g., a -2\,mm deformation means they are separated by 2\,mm. Prior to contact, thinner caps exhibit lower audio amplitudes and are closer to the open boundary condition than thicker caps. However all cap thicknesses demonstrate a downward deviation in audio amplitude centered between 0 and 2\,mm of contact. 
In Fig. \ref{fig:tihj}C, the resonant frequency $f$ is plotted as a function of force $F$. All the caps produce nonmonotonic boundary condition transitions, while the thinner caps show greater frequency fluctuations during the boundary condition change. 
The thinnest cap signal saturates at around 3\,N of load, where the cap is compressed against the rigid resonant tube and can deform no more. The force-frequency slopes are consistent with cap stiffness trends.

%\clearpage

%%%%%%%%%%%%%%%%%%%%%%%%%%%%%%%%%%%%%%%%%%%%%%%%
\subsection{Cap design feature: Hole, $h$}

Fig. \ref{fig:tihj}D shows the audio amplitude with varying cap wall thickness in designs with a 3\,mm diameter hole (t1h3-t5-h3). As compared with the cases without a hole, all taxels have a lower amplitude when there is no contact. Upon contact, varied deviations in the amplitude arise. However, all signals increase and remain above the unloaded amplitude with enough deflection. This amplitude cross-over threshold occurs for the smallest deflection for the thinnest cap, which also has the lowest amplitude prior to initial contact. 
In Fig. \ref{fig:tihj}E, the force-frequency relationship is plotted while employing the amplitude crossover threshold to produce monotonic results. The shape of the force sensing curves is similar to those found in Fig. \ref{fig:tihj}C, except for the omission of data at the lowest force levels. Table \ref{tab:tih3} reports the each force taxel's resulting approximate linear ranges and sensitivities. The lower bound of force range is from the minimum detectable force after amplitude thresholding. Visual inspection dictates the upper bound as where the signal begins to saturate. 

\begin{table}[h]
\centering
\begin{threeparttable}
\caption{Amplitude threshold value, force range and sensitivity for caps with 3mm hole, plotted in Fig. \ref{fig:tihj}E. 
\label{tab:tih3}}
\setlength{\tabcolsep}{12pt}%
\begin{tabular}{@{}lllll}
\toprule
$t$ [mm] & Threshold [AU] & F$_\text{min}$ [N]  & F$_\text{max}$ [N] & Sensitivity [Hz/N]  \\
\midrule
 1 & 0.17  & 0.5      &  2.6    &   19   \\
2 &  0.15 & 0.6        &   6.1   &   8.8    \\
3 &  0.12  &  0.5      &   10.   &   5.1  \\
4 &   0.13  & 1.3     &   15   &   3.6  \\
5 &   0.15  & 1.5      &   15   &   3.2  \\
\bottomrule
\end{tabular}
\end{threeparttable}
\end{table}

In Fig. \ref{fig:tihj}F, amplitude is plotted across caps with the same thickness, 3\,mm, but with varying hole size (t3h0-t3h5). Larger holed caps exhibit lower sound amplitude prior to contact, which is expected as they are more physically similar to an open tube. 
As a result, the larger hole allows for a lower amplitude threshold. As seen in the monotonic force-frequency curves in Fig. \ref{fig:tihj}G and tabulated values in Table \ref{tab:t3hj}, 
caps with larger holes capture smaller forces in their sensing range when using amplitude thresholds. With the exception of the t3h5 calibration curve, for which the hole is a substantial size compared with the cap and lower forces are measured, these sensitivities are within 8\% of each other. 

\begin{table}[h]
\centering
\begin{threeparttable}
\caption{Amplitude threshold value, force range and sensitivity for caps with 3mm wall thickness, plotted in Fig. \ref{fig:tihj}G.
\label{tab:t3hj}}
\setlength{\tabcolsep}{12pt}%
\begin{tabular}{@{}llll}
\toprule
$h$ [mm] & Threshold [AU] & F$_\text{min}$ [N]  & Sensitivity [Hz/N]  \\
\midrule
 0 & 0.22  & 1.6   &   5.1   \\
1 &  0.19 &   1.1  &   4.9     \\
3 &  0.14  &   0.9   &  5.3    \\
5 &   0.06  &   0.2  &  7.9   \\
\bottomrule
\end{tabular}
\end{threeparttable}
\end{table}

\begin{figure}[h]
\centering
	\includegraphics[width=0.9\linewidth]{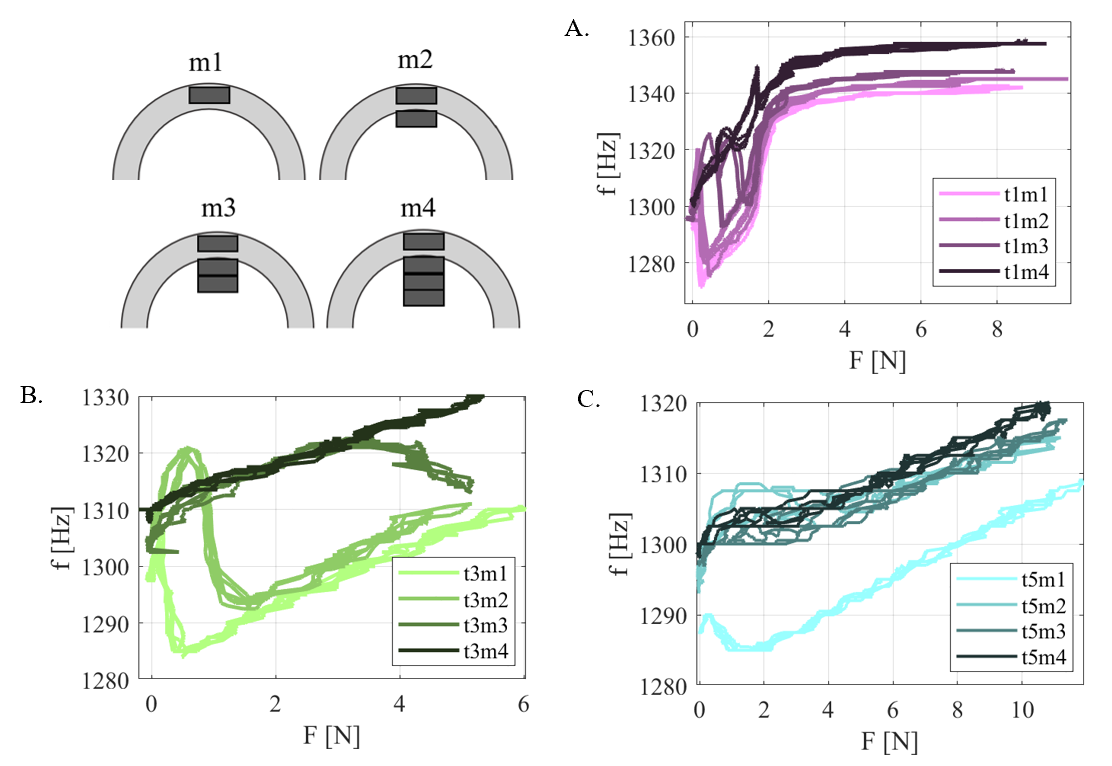}
	\caption{Characterization of added mass $m$, as shown in the schematic legend of stacked magnets used as mass on cap. Frequency $f$ versus force $F$ for wall thickness for A. 1\,mm, B. 3\,mm, and C. 5\,mm caps. Darker lines indicates higher masses. 
	}
	\label{fig:timk}
\end{figure}

%%%%%%%%%%%%%%%%%%%%%%%%%%%%%%%%%%%%%%%%%%%%%%%%%%%%%%%5
\subsection{Cap design feature: Mass, $m$}

Fig. \ref{fig:timk} depicts how magnets attached to the tip of the cap alter force-frequency relationships in the transition region, where A, B, and C show the results for the 1, 3, and 5\,mm cap thicknesses, respectively. The added mass varies from 50\,g (m1) to 200\,g (m4), with darker lines indicating higher mass. 
Overall, increasing mass mitigates the frequency fluctuations from the boundary condition transition. We observe nonmonotonic behavior in the tested force range with 50\,mg added mass in all the three cap thicknesses. This boundary condition effect resolves with 200\,mg for 1\,mm thickness and 3\,mm thicknesses, and 50\,g for 5\,mm thickness caps. As expected, the thinner caps require more mass to remove the boundary condition transition. 
For both the 1 and 3\,mm thicknesses, intermediate levels of mass shifts the minima frequency locations to higher force levels. The mechanism that causes the t3m3 force-frequency curve to have this upward then downward sloping shape is unclear; the decrease is out of the typical force range for a boundary condition transition and perhaps due to the mechanics of the added mass on the cap. The magnets have nonzero volume and are rigid, which may account for the hysteretic effect observed within some transition regions, e.g., in t1m3.

%%%%%%%%%%%%%%%%%%%%%%%%%%%%%%%%%%%%%%%%%%
\section{AcousTac demonstrations} 
\label{sec:demo}

\subsection{Test systems}

\begin{figure}[h]
    \centering \includegraphics[width= 1\linewidth]{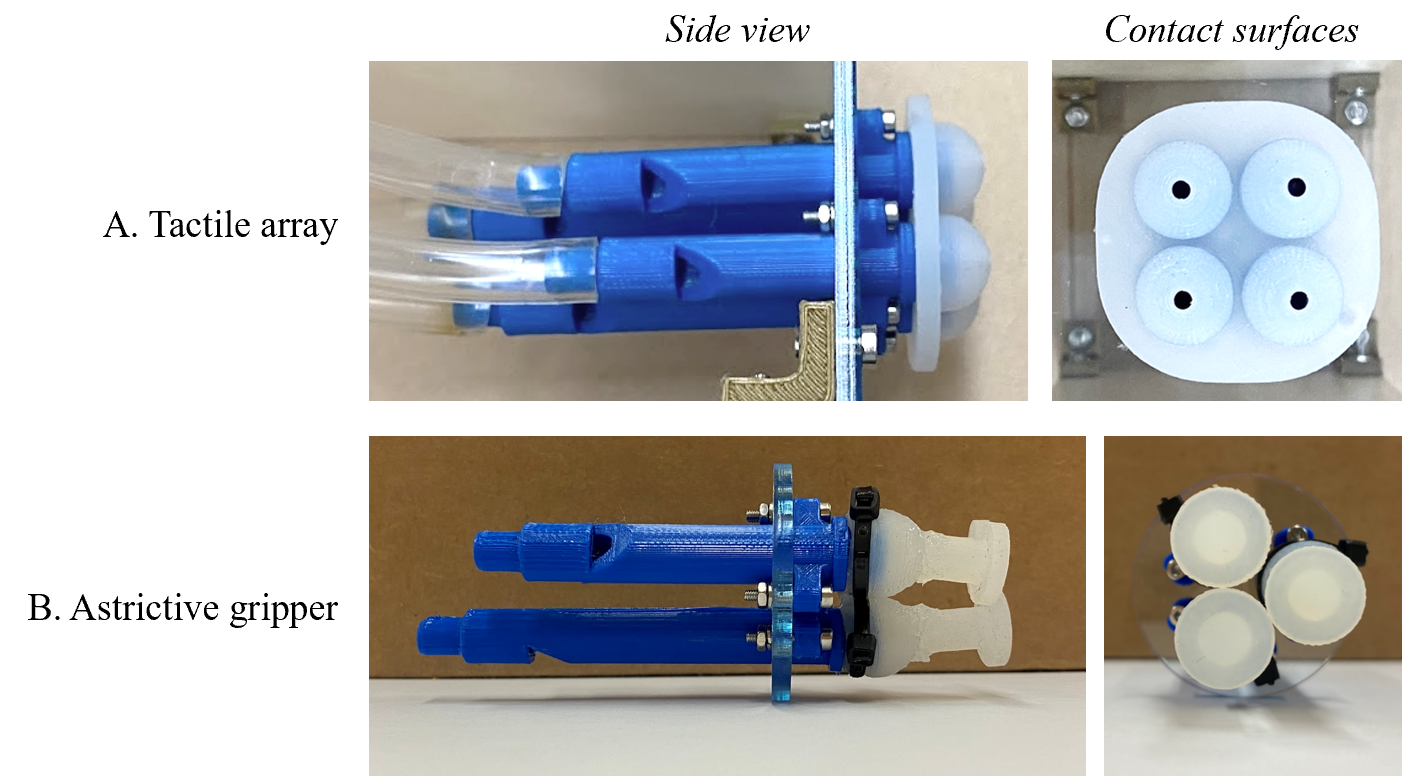}
    \caption{A. Tactile array with four taxels of length L = 41, 47, 59, 65\,mm. Caps have 3\,mm wall thickness and a 3\,mm hole diameter (t3h3). B. Astrictive gripper with three taxels of length L = 41, 47, 59\,mm. Caps have 2.5mm wall thickness and suction cups adhered to the tops. 
    }
    \label{fig:demoPics}
    %\vspace{-5mm}
\end{figure}

Two multi-taxel systems are shown in Fig. \ref{fig:demoPics}: (A) a tactile array and (B) an astrictive gripper. 

First, we combine four taxels of different lengths for an indicied, force and contact sensitive 2x2 array. The taxel lengths are consistent with those in Sec. \ref{sec:results:tubelength}: 41, 47, 59, and 65\,mm. We position the four taxels in a grid pattern, spaced 15\,mm apart center-to-center. 
The end-caps have a 3\,mm wall thickness and 3\,mm hole (t3h3). 
For the processing of the signal from the L59 taxel, we use the sensitivity value from Table \ref{tab:tih3} to translate frequency to force. For the three other taxel lengths, we scale the force by a ratio of $L^2$ according to Eqn. \ref{eqn:Delta_fL}. All taxels receive air from the same source. As varying flow rate produces slight differences in each taxel's frequency, and can change as neighboring taxel holes get covered, the measured frequency at light contact is set as the unloaded frequency as an initialization calibration procedure.

Next, we integrate three taxels into a force sensitive astrictive gripper. Inspired by prior work \cite{huh_2021, kang_adhesivesreview_2021}, we implement passive suction cups to grip onto objects. 
The taxels are evenly spaced 15.9\,mm apart, with rigid tube lengths of 41, 47, and 59\,mm.  The caps have 2.5\,mm wall thickness, and we adhere a suction cup to each cap using an air-cured silicone glue (Smooth-On Sil-Poxy). The suction cups are similarly cast from silicone (Smooth-On Dragon Skin 30) and are 13\,mm in diameter and 2.5\,mm thick. The suction cups sit on posts 7\,mm in diameter and 11\,mm tall. 
We affix the caps to the rigid tube and wet the suction surface with tap water to increase suction force. 
Because the introduction of suction cup structures to the caps alters the frequency response of this system, we report frequency and not force. 

\begin{figure}[h!]
    \centering \includegraphics[width= 1\linewidth]{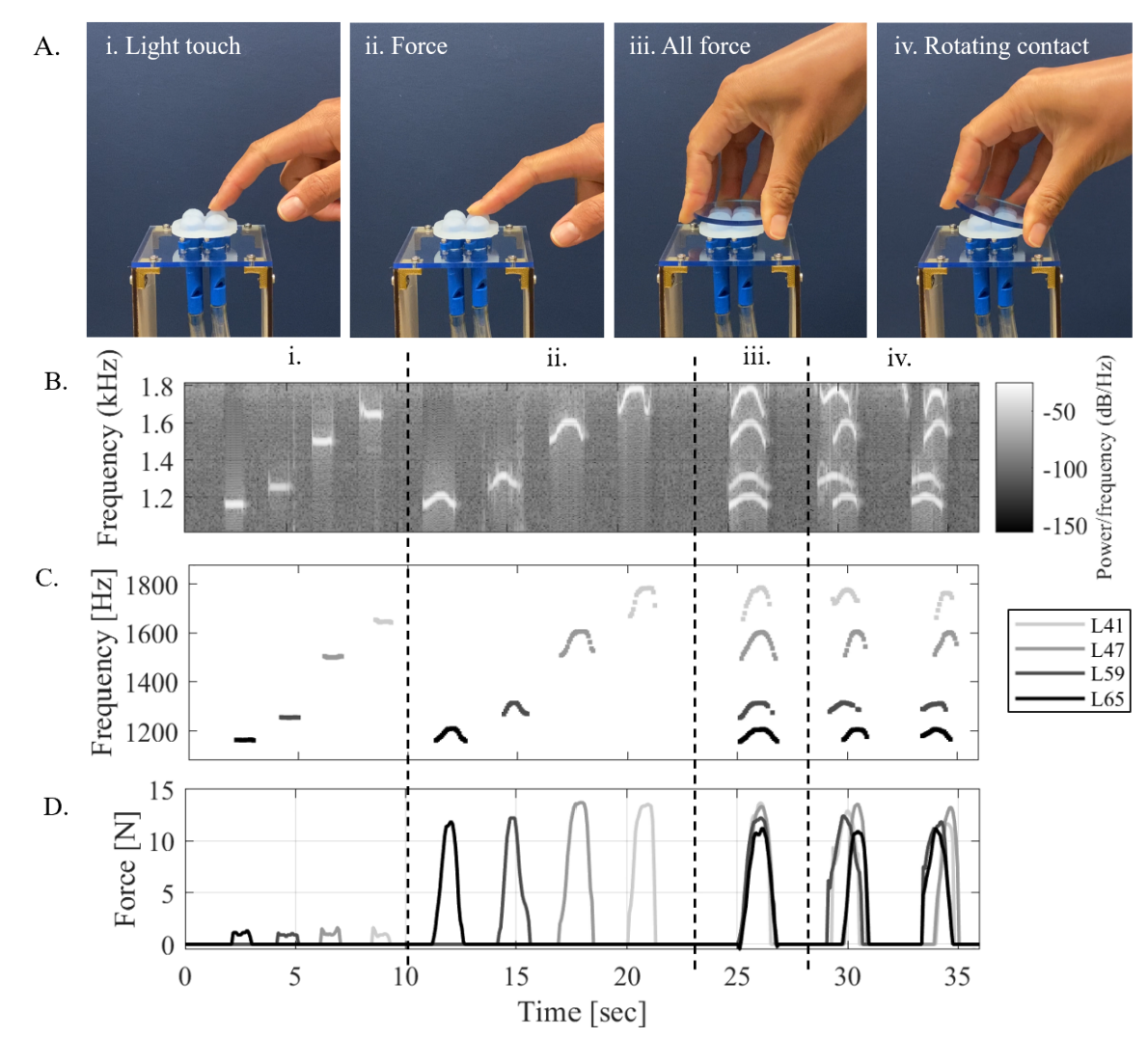}
    \caption{Robotic demonstration of tactile array with caps t3h3. We palpate taxels individually, with a light touch (i) and then with a forceful touch (ii). We then use an acrylic round to press all the taxels down simultaneously (iii). Then we use the same round and roll/rotate across the taxels, across both directions (iv). A. Photos corresponding to the sequences. B. Spectrogram. C. Signal processed with amplitude and frequency over time. D. Normal force over time. 
   }
    \label{fig:demo1}
\end{figure}

\begin{figure}[h!]
    \centering \includegraphics[width= 1\linewidth]{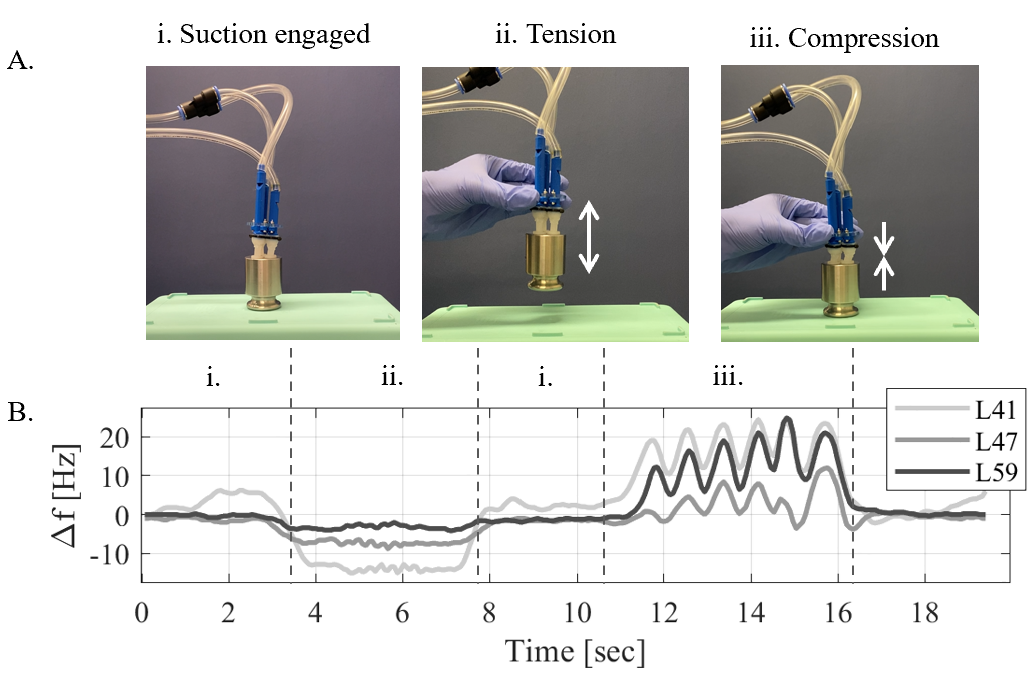}
    \caption{Demonstration of suction cup gripper with three AcousTac taxels tested on a 500\,g weight. All three suction cups are engaged (i), and exhibiting some load due to air hose routing. Gripper is lifted along with the weight, and then hefted so the weight bounces up and down (ii). Weight is set back on the surface and compression forces are applied (iii). A. Photos from demonstration. B. Net frequency change of the three taxels over time.
   }
    \label{fig:demo2}
\end{figure}

\subsection{Tactile array interaction}

Shown in Fig. \ref{fig:demo1}A, we contact and palpate the caps individually with one finger and together with a piece of acrylic over four different sequences. First (i), we make light contact with each taxel individually, started from L65 and ending with L41. In the same order, we push down on each taxel individually with greater force (ii). In the third sequence (iii), an acrylic plate is pushed down on all the taxels simultaneously. In the last sequence (iv), the same acrylic piece applies a rolling forceful contact across the array in two orientations rotated by 90 degrees, side to side and front to back. 
Fig. \ref{fig:demo1}B shows the audiospectrogram in the relevant frequency range. We extract the peak frequencies from the spectrogram (Fig. \ref{fig:demo1}C) and translate them to force (Fig. \ref{fig:demo1}D). 
Signals are as expected. With rolling contact, we observe two taxels loaded prior to the other two, showing the different between the two rolling orientations. 
We use a linear mapping enabled by the addition of a hole and therefore monotonic force-frequency relationship. This tactile array demonstrates the practicality of holes in the caps, exhibiting low amplitude of sound when not in contact, and resonance when loaded.

%%%%%%%%%%%%%%%%%%%%%%%%%%%%%%%%%%%%%%%%%%
\subsection{Astrictive gripping}
Shown in Fig. \ref{fig:demo2}A, we manipulate a 500\,g weight, imparting tension and compression via suction cup gripping. Fig. \ref{fig:demo2}B shows the net frequency of the three taxels. In the demonstration, the suction cups adhere to the smooth, flat surface of the weight (i). The gripper lifts the weight and hefts repeatedly with weight still attached (ii), varying the tension loading. In part (iii), we compress the suction cups against the weight. The resonant frequency signals show expected trends, in which 
the sensor captures the oscillating tension forces at approximately 4\,Hz in (ii) and the larger compression force at about 1\,Hz in (iii). 

%%%%%%%%%%%%%%%%%%%%%%%%%%%%%%%%%%%%%%%%%%
\section{Discussion}
\label{sec:disc}
The experimental results show the substantial effects of tube and cap geometry on the sound signals generated by AcousTac. 
Parameters including tube length and cap wall thickness affect frequency and force measurements as expected from simple first-principle design models. 
For example, as the tube length decreases, a given deformation results in a higher frequency output range. So, shorter tubes have higher sensitivities and also higher total frequencies. We leverage this length-frequency relationship in order to read multiple force signals simultaneously by assigning different tube lengths to different taxels in system demonstrations. Designers can further tune the cap wall thickness or material stiffness for a desired force sensing range. The fabrication of these taxels is simple, requiring only a 3D printer and silicone casting equipment. In part, the simplicity of this system is because there is no need for the integration of precise of sensitive electronics into the skin. 

We observe a change in boundary conditions when contact occurs with the soft taxel cap. The compliant material produces a boundary condition that lies in between the open and closed cases. We posit that this damping results in the nonmonotonic force-frequency relationship that occurs upon initial touch during experiments. 
In order to tackle this issue, the addition of a hole or mass at the tip of the cap enables monotonic force readings, not otherwise achieved. 
When using holes, rather than mass, the sensor is generally quieter when no contact is being made, which is advantageous if operating near humans who may not want to hear the sensors at all times. The larger the hole, the quieter the sensor when not in contact. 
The presence of the hole also enables pre-touch and close proximity sensing; prior to contact, at distances of approximately 1\,mm, a unique set of amplitude and resonant frequency signals are generated. 
At the same time, adding mass is a useful solution if a hole is not an appropriate option, for example when handling objects in wet or dirty environments where matter would enter the hole. In addition, the taxels with holes only produce reliable sounds when the hole becomes sealed by contact, which may not always occur. 

Throughout this work, we use a common smartphone microphone to measure signals, demonstrating the accessibility and non-specialized requirements to implement AcousTac. In the current implementation, we implement audio processing parameters that dictate the temporal and force resolution. While this is not a physical limitation of the taxel sound emission, sound collection hardware and processing speeds will influence overall system performance and the ability to detect fast transient loads. We operated experimental trials in a normal laboratory environment. In environments with loud or complex ambient noise, more sophisticated processing methods may be needed. 

\subsection{Limitations and future work}
Spatial resolution is not the focus of this current work and could be improved in future work. 
Taxels need to be spaced far enough apart such that a deformed cap does not interfere with the neighboring taxel caps. Therefore, as taxel diameters get smaller, they can be packed more closely. As taxels reduce in size, output frequency, air supply, fabrication precision, and microphone requirements may need to change. Another issue could arise where the surrounding taxels interfere with the airflow near the edge-orifice. If the edge-orifice area of the tube is obstructed, resonance is inhibited, though we do not observe any issues for taxels spaced 15\,mm apart. 
For sensors at substantially different length scales, a resonance generating geometry other than the current edge-orifice may be more appropriate from a resonance, packing, and physical implementation standpoint. With the addition of more taxels, future work should also investigate the limit on how many simultaneous frequencies can be produced within unique frequency ranges. 

Throughout this work, we assume superposition holds for sounds signals from systems of multiple taxels. This seems to be true for the current design, where each taxel operates the same independently of the other taxels. However, with closely overlapping frequency bands, we observe heterodyning and difficulty processing the signals as a result. We also observe a dependence on microphone location due to constructive and destructive signal interference. 
Each taxel is separated by rigid materials in this study. When resonating soft structures are physically coupled together, their resonant frequencies can interact in complex and nonlinear ways \cite{librandi_2021}, a topic to further investigate in fully-soft resonating structures.

We characterized AcousTac under normal loading conditions, yet oblique loading occurs in the real robot manipulation. As a preliminary test, we sheared caps with and without a hole under 5N normal force, with up to 2N lateral force. We did not find the sliding and shear forces to affect the normal force measurements. By modifying the cap geometry, AcousTac structures could potentially be specialized for other particular types of loading and deformation measurements in future work. 

AcousTac uses the ambient atmosphere to transmit information. The atmosphere must be gaseous, as this resonance generation utilizes the compressible nature of the ambient air. The atmosphere also must be dense enough such that the sound propagates, i.e. not the vacuum of space. Extreme changes in temperature, density, and humidity would require additional calibration steps as they alter resonance and sound propagation \cite{maurice_2022}. As an open fluid system, the robot also needs to be able to provide a continuous flow of air when activating the sensors. 
Pneumatic systems for locomotion in untethered robots \cite{drotman_elecfree_2021} and robots in extreme environments \cite{mahon_soroee_2019} could potentially power AcousTac for sensing on mobile robots. 
One relatively untapped benefit of active acoustic tactile sensing is that sound can be measured across large distances, even without physical contact with the resonating system. Thus, if loud enough, information about robot contact state is detected by any microphone within audible range. 

%%%%%%%%%%%%%%%%%%%%%%%%%%%%%%%%%%%%%%%%%%
\section{Conclusion}
\label{sec:concl}
In this work, we present AcousTac, a pneumatic-driven electronic-free tactile sensor that emits deformation and force information through active acoustic resonance. AcousTac is capable of producing monotonic force measurements without hysteresis, captured by a remote microphone. Because there are no electronics located at the skin, the taxels are simple and cheap to fabricate and resilient to loading that might otherwise fatigue or break electrical wires or connectors.
By characterizing a range of sensor parameters and comparing to theory, this work serves to guide future implementations of AcousTac across a range of force scales and applications. 
For example, it is an attractive option for operation where electronics cannot, e.g., in an Magnetic Resonance Imaging (MRI)  machine or in harsh environmental conditions, e.g., when handling volatiles. AcousTac could also potentially be utilized for wireless communication between different agents, where any agent with a microphone can sense the state of another. 
Ultimately, sound holds great potential to enable soft robots to communicate information across distances without wires; with soft structures generating resonant sound, emitted frequencies sensitive to robot state can be harnessed with audio transducers that are already integrated in robotic systems. %Thus tactile sensing becomes an integrated soft and smart robot structure design challenge. 

\section*{Acknowledgments}
The work of M. Li is supported by the National Aeronautics and Space Administration grant No.80NSSC20K1166 through a Space Technology Research Fellowship. We acknowledge the assistance of Jadesola Aderibigbe who made the suction cups used in the demonstration, Jungpyo Lee and Sebastian Lee who assisted with robot arm control, and Christopher Yahnker and Tae Myung Huh who met with the authors to give their perspective.

\nocite*{}
% \bibliographystyle{unsrt}
 % \bibliography{soro}

\end{document}